\documentclass[10pt,twocolumn,letterpaper]{article}

\usepackage[pagenumbers]{cvpr} 

\usepackage{graphicx}
\usepackage{amsmath}
\usepackage{amssymb}
\usepackage{booktabs}

\usepackage[pagebackref,breaklinks,colorlinks]{hyperref}

\usepackage[capitalize]{cleveref}
\crefname{section}{Sec.}{Secs.}
\Crefname{section}{Section}{Sections}
\Crefname{table}{Table}{Tables}
\crefname{table}{Tab.}{Tabs.}

\def\cvprPaperID{9426} 
\def\confName{CVPR}
\def\confYear{2023}

\usepackage{multirow}

\newcommand{\dataset}{WASD}
\newcommand{\github}{https://github.com/Tiago-Roxo/WASD}

\begin{document}

\title{WASD: A Wilder Active Speaker Detection Dataset}


\author{Tiago Roxo,~~~Joana C. Costa,~~~Pedro R. M. Inácio,~~~Hugo Proença\\
IT - Instituto de Telecomunicações \\
University of Beira Interior, Portugal\\
{\tt\small \{tiago.roxo, joana.cabral.costa\}@ubi.pt, \{prmi, hugomcp\}@di.ubi.pt}
}

\maketitle



\begin{abstract}

Current Active Speaker Detection (ASD) models achieve great results on AVA-ActiveSpeaker (AVA), using only sound and facial features. Although this approach is applicable in movie setups (AVA), it is not suited for less constrained conditions. To demonstrate this limitation, we propose a Wilder Active Speaker Detection (\dataset) dataset, with increased difficulty by targeting the two key components of current ASD: audio and face. Grouped into 5 categories, ranging from optimal conditions to surveillance settings, \dataset~contains incremental challenges for ASD with tactical impairment of audio and face data. We select state-of-the-art models and assess their performance in two groups of \dataset: Easy (cooperative settings) and Hard (audio and/or face are specifically degraded). The results show that: 1) AVA trained models maintain a state-of-the-art performance in \dataset~Easy group, while underperforming in the Hard one, showing the 2) similarity between AVA and Easy data; and 3) training in \dataset~does not improve models performance to AVA levels, particularly for audio impairment and surveillance settings. This shows that AVA does not prepare models for wild ASD and current approaches are subpar to deal with such conditions. The proposed dataset also contains body data annotations to provide a new source for ASD, and is available at \url{\github}.
    
\end{abstract}


\section{Introduction}
\label{sec:intro}

Active Speaker Detection (ASD) aims to identify, from a set of potential candidates, active speakers on a given visual scene~\cite{roth2020ava}. Currently, this assessment is done at the video frame level based on facial cues and sound information. Despite its application in several topics such as speaker diarization~\cite{gebru2017audio, chung2019said, chung2020spot}, human-robot interaction, or speaker tracking~\cite{qian2021audio, qian2021multi}, its applicability in wild conditions is still an open issue.

\begin{figure}[t]
  \centering
  \includegraphics[width=0.5\textwidth]{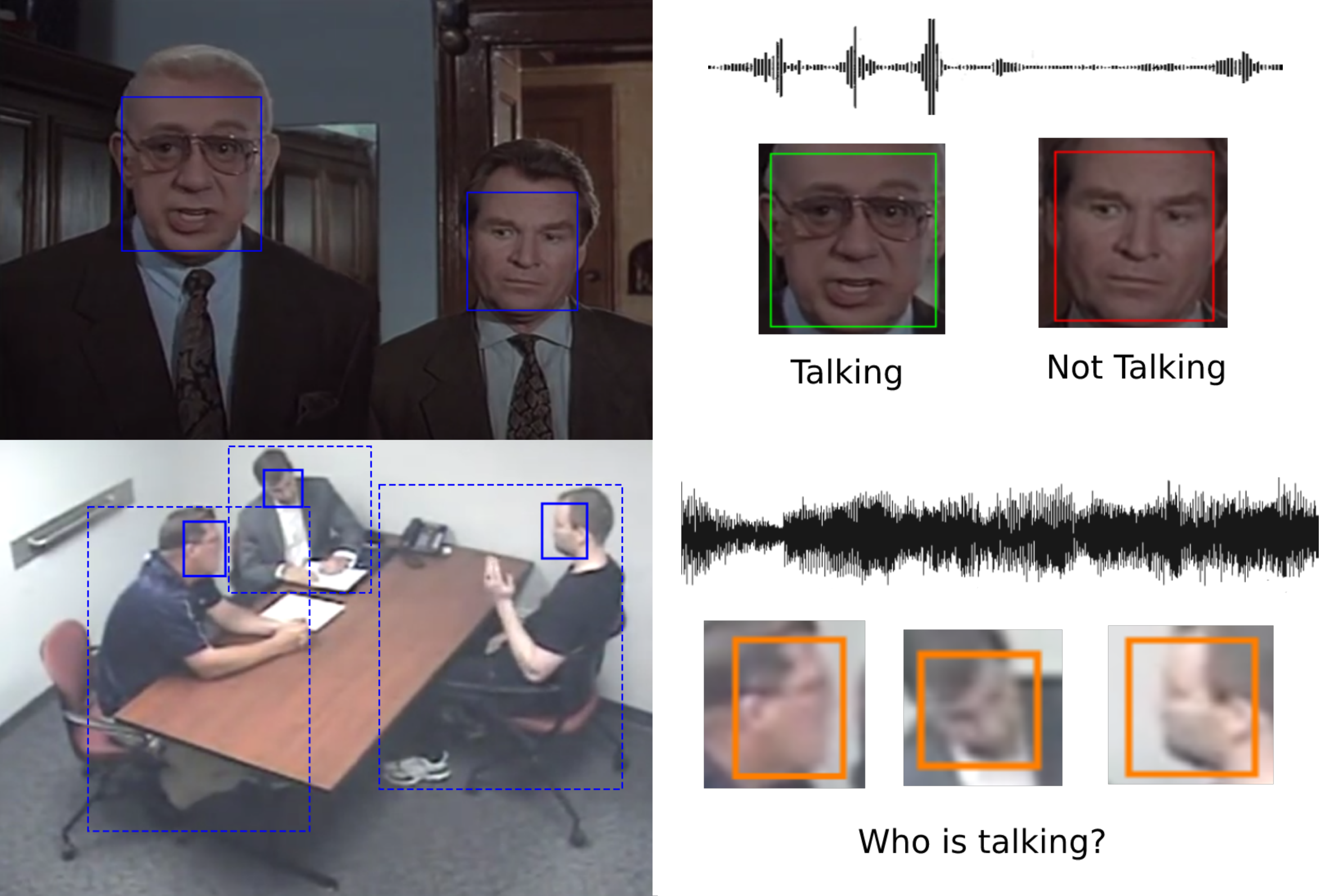}

   \caption{AVA-ActiveSpeaker state-of-the-art models achieve over 94\% mean Average Precision (mAP) in active speaker detection, solely based on \textbf{face} and \textbf{audio data}. However, this approach may not be suited for uncooperative poses, non-guaranteed face access, or unreliable image/audio quality. How well do these models perform in such scenarios? And can body information aid in this task?}
   \label{fig:init_image}
\end{figure}

\begin{table*}[!tb]
    \centering
    \small
    \renewcommand{\arraystretch}{1.05}
    \caption{Feature comparison of ASD datasets. AVA-ActiveSpeaker is represented as AVA. If datasets contain information regarding a feature, its absence is presented with $\times$, while its presence with $\checkmark$. \dataset~has a high number of hours, with increased number of faces and reduced face tracks (culminating in higher average video duration), Frames Per Second (FPS) variability, and increased talking percentage. The most disciminative factors are demographic representation, surveillance conditions, and body data annotations.}
    \begin{tabular}{c*{10}{c}}\hline
    
        \multirow{2}{*}{\makebox[5em]{\textbf{Dataset}}} & Total & Number of & Face & Video & FPS & \multirow{2}{*}{\makebox[3em]{Talking \%}}  & \makebox[3em]{Demographic} & \makebox[3em]{Surveillance} & Body  \\
        & Hours & \makebox[4em]{Faces (M)} & \makebox[4em]{Tracks (m)} & \makebox[4em]{Duration (s)} & \makebox[4em]{Variability} & & Representation & Conditions & Data \\  
        \hline\hline

        Columbia \cite{chakravarty2016cross} &
        1.5 & 0.2 & - & - & $\times$ & - & $\times$ & $\times$ & $\checkmark$ \\

        Talkies \cite{alcazar2021maas} &
        4.2 & 0.8 & 23.5 & 1.5 & - & - & - & $\times$ & $\times$ \\

        EasyCom \cite{donley2021easycom} &
        6.0 & - & - & - & $\times$ & - & $\times$ & $\times$ & $\times$ \\

        ASW \cite{kim2021look} & 
        30.9  & - & 11.5 & $\sim$10 & - & 57.9 & - & $\times$ & $\times$\\

         AVA \cite{roth2020ava}  & 
        37.9 & 3.7 & 38.5 & $\leq$10 & $\checkmark$  & 24.2 & - & $\times$ & $\times$\\

        \hline
        \textbf{\dataset} &
        30.0 & 7.4 & 9.8 & $\sim$28 & $\checkmark$ & 84.6 & $\checkmark$ & $\checkmark$   & $\checkmark$ \\
    
        \hline
        
    \end{tabular}
    \label{table:dataset-soa}
\end{table*}

The state-of-the-art dataset for ASD is AVA-ActiveSpeaker~\cite{roth2020ava}, composed of several Hollywood movies, with diversity in languages, recording conditions, and speaker demographics, totalling in 38 hours and over 3 million face images. Although AVA-ActiveSpeaker has some challenging aspects, it still is not a perfect representation of \textit{in-the-wild} data~\cite{roth2020ava}, since it assesses ASD in movies, a setup with controlled (scripted) action and speaking, with adequate audio and image quality. This motivates state-of-the-art models to identify active speakers solely based on audio and face data, disregarding other informations such as speaking context or body expressions. This is particularly problematic since ASD in wild conditions can not assume face availability, subject cooperation, and good audio quality, as shown in Figure~\ref{fig:init_image}. To overcome these limitations, we propose a Wilder Active Speaker Detection (\dataset) Dataset.

\dataset~aims to preserve the challenging characteristics of AVA-ActiveSpeaker while increasing the difficulty of ASD by targeting the two key components state-of-the-art models use: face and audio. 
We select videos from YouTube and group them into 5 categories, based on a set of features targeted at face and audio impairment. The categories range from optimal conditions (face availability and good audio quality), to surveillance settings (non-guaranteed face access, subject cooperation, or sound quality). The increasing scale of ASD challenges can be useful for: 1) assess the ability of current models to deal with wild conditions and specific aspect impairment (audio, face, or a combination of both); 2) evaluate the limitations of AVA-ActiveSpeaker to prepare models for wild conditions; and 3) show the limitations of face and audio dependency for wild ASD, easing the identification of model improvements towards this goal. By selecting YouTube videos from real interactions, \dataset~also contains expressions, sudden interruptions, and interactions that movies hardly contain. These additional challenges, enhanced by the variability of demographics in \dataset, contribute to a challenging ASD dataset where state-of-the-art models can not easily perform. Furthermore, \dataset~provides body data annotations to motivate the development of models using body information to complement face and audio data in (wild) ASD. To summarize, the main contributions are:

\begin{itemize}
    \item We propose \dataset, a ASD dataset divided into 5 categories with incremental ASD challenges, targeting audio quality and face availability, ranging from optimal conditions to surveillance settings;
    \item We assess and show the limitations of AVA-ActiveSpeaker training and state-of-the-art approaches for ASD in setups with audio impairment, facial occlusion, and surveillance settings.
\end{itemize}

\begin{figure*}[t]
  \centering
   \includegraphics[width=0.99\linewidth]{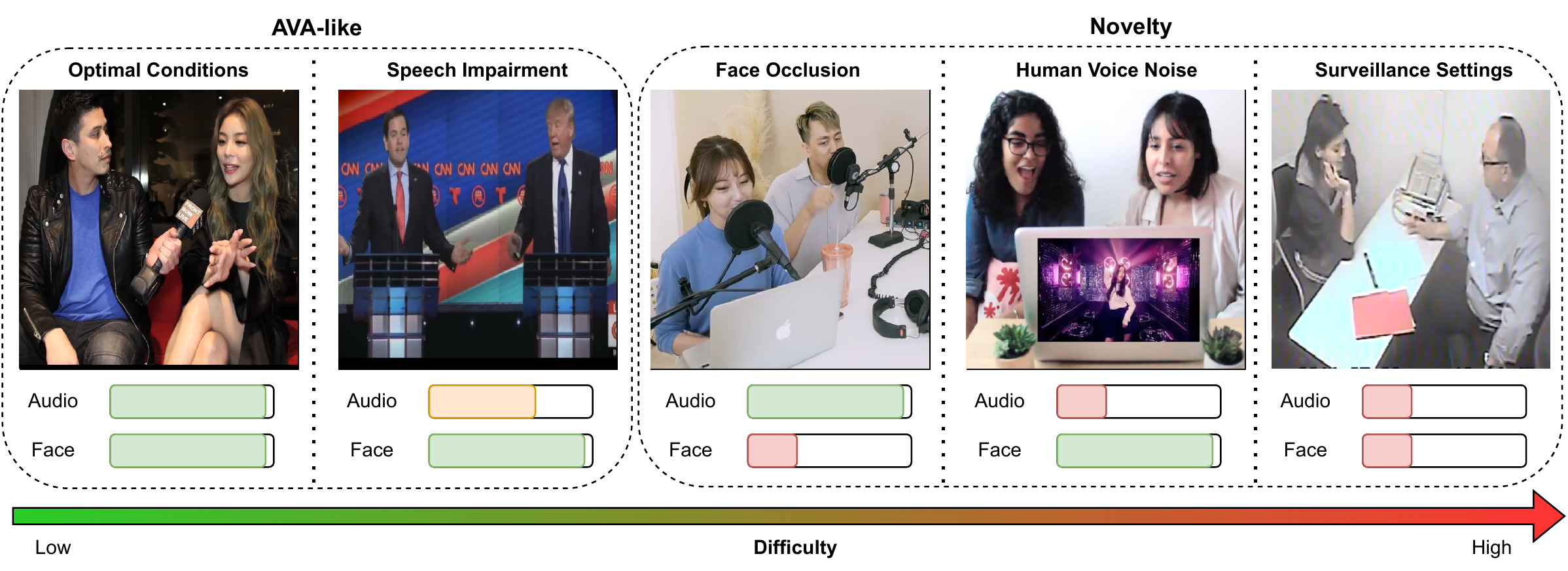}
   \caption{Considered categories of \dataset, with relative audio and face quality represented. Categories range from low (Optimal Conditions) to high (Surveillance Settings) ASD difficulty by varying audio and face quality. Easier categories contain similar characteristics to AVA-ActiveSpeaker (AVA-like), while harder ones are the novelty of \dataset.}
   \label{fig:categories_dataset}
\end{figure*}

\section{Related Work}
\label{sec:related-work}

\textbf{Active Speaker Detection.}
Works on ASD have evolved from facial visual cues~\cite{saenko2005visual, everingham2009taking, patrona2016visual} to audio as primary source~\cite{chakravarty2015s, ding2020personal}, to multi-modal data combination~\cite{roth2020ava, alcazar2020active, tao2021someone, alcazar2021maas, kopuklu2021design}. Since the introduction of AVA-ActiveSpeaker~\cite{roth2020ava}, combining audio with facial features is the \textit{de facto} way to predict active speakers.
Large 3D architectures~\cite{chung2019naver}, hybrid 2D-3D models~\cite{zhang2019multi}, and large-scale pretraining~\cite{chung2016out, chung2019perfect} for audio-visual combination are amongst some of the following works. Despite the viability of these approaches, feature embedding improvement~\cite{hadsell2006dimensionality} or attention approaches~\cite{vaswani2017attention, afouras2020self, cheng2020look} were necessary to improve ASD. Creating two-step models, where the first focuses on short-term analysis (audio with face combination) and the second on multi-speaker analysis, is the approach from various recent works~\cite{alcazar2020active, kopuklu2021design, zhang2021unicon, alcazar2021maas}. ASC~\cite{alcazar2020active} focused on long-term multi-speaker analysis via temporal refinement, ASDNet~\cite{kopuklu2021design} used a similar approach for inter-speaker relations, with improved visual backbones, and UniCon~\cite{zhang2021unicon} relied on audio-visual relational contexts with various backbones. Improving speaker relation representation via Graph Convolutional Networks (GCN)~\cite{welling2016semi} is also a viable approach to assess context information~\cite{alcazar2021maas, min2022learning}. Diverging from two-step training, end-to-end models have also emerged for ASD~\cite{tao2021someone, alcazar2022end, min2022learning}. TalkNet~\cite{tao2021someone} focused on improving long-term temporal context with audio-visual synchronization, while EASEE~\cite{alcazar2022end} included GCN to complement spatial and temporal speaker relations.

\textbf{Datasets.} There is a variety of available datasets suited for ASD, such as frontal speaker data, designed for speech recognition~\cite{hazen2004segment, patterson2002cuave}, voice activity detection~\cite{tao2017bimodal}, and diarization~\cite{gebru2017audio} datasets. However, these are limited in subject diversity and talking scenarios, diminishing their relevance. With increased talking variability, datasets derived from movies and TV shows have also been reported~\cite{giraudel2012repere, ren2016look, everingham2006hello, hu2015deep}, limited by the low number of annotated hours. Other setups related with ASD are lip reading datasets~\cite{son2017lip, chung2016lip, chung2017lip, afouras2018lrs3, Nagrani2017, Chung2018}, whose purpose diverges from ASD since their goal is to infer the words pronounced from a given speaker. Recently there is a greater focus on specific ASD datasets~\cite{chakravarty2016cross, alcazar2021maas, donley2021easycom, roth2020ava, kim2021look}, whose task is to determine the talking speaker from a set of admissible candidates. Columbia~\cite{chakravarty2016cross} contains 87 minutes of a panel discussion, with up to 3 visible speakers. Talkies~\cite{alcazar2021maas} focuses on low duration videos, totalling 4 hours, with an average of 2.3 speakers and off-screen speaking. Easycom~\cite{donley2021easycom} is designed for multiple tasks related with augmented reality, composed of various sessions of speakers sat at a table, with background noise. AVA-ActiveSpeaker~\cite{roth2020ava} is the state-of-the-art dataset, with over 150 Hollywood videos, totalling almost 38 hours, with demographic diversity and dubbed dialogues. ASW~\cite{kim2021look} was proposed with over 30 hours, from 212 videos randomly selected from the VoxConverse~\cite{Chung2020}, containing various sets of interviews. The proposed dataset, \dataset, brings challenging sets, \textit{in-the-wild} videos, demographic diversity, and body data annotations. The main characteristics of our dataset relative to others are presented in Table~\ref{table:dataset-soa}.

\begin{figure*}[!tb]
    \centering
    \begin{subfigure}[b]{0.33\textwidth}
         \centering
         \includegraphics[width=\textwidth]{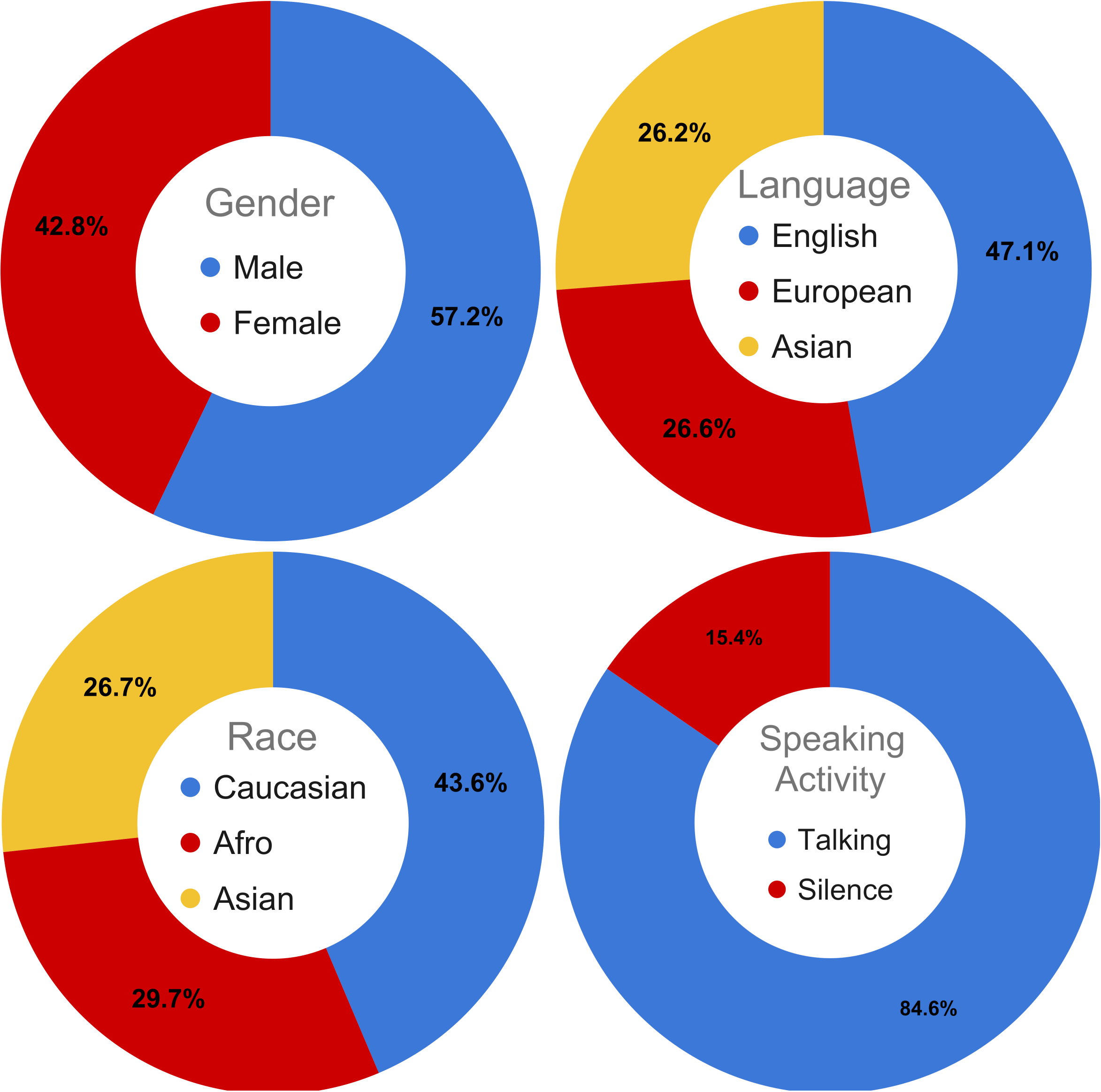}
     \end{subfigure}%
     \begin{subfigure}[b]{0.24\textwidth}
         \centering
         \includegraphics[width=\textwidth]{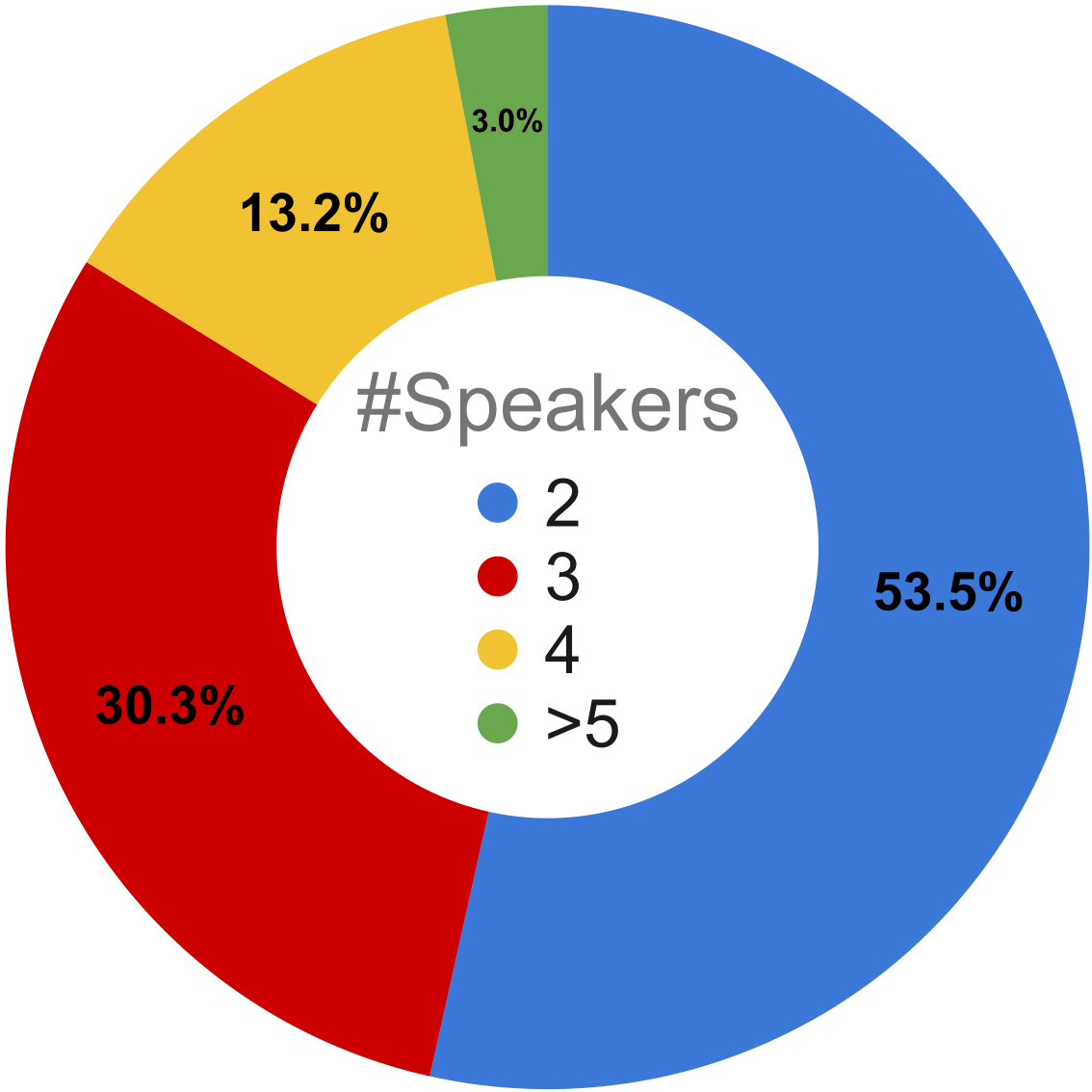}
         \label{fig:number_speakers}
     \end{subfigure}%
     \begin{subfigure}[b]{0.42\textwidth}
         \centering
         \includegraphics[width=\textwidth]{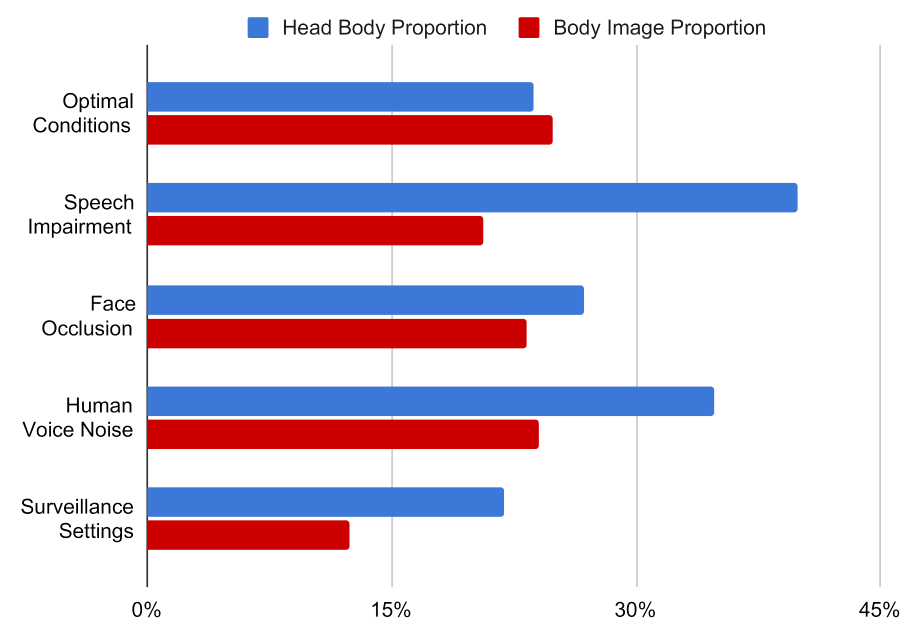}
         \label{fig:Head_Body_Proportion_Small}
     \end{subfigure}%
    \caption{Gender, language, race, speaking activity, and number of speakers distribution of \dataset. Afro refers to African and Afro-American people. On the right, distribution of head-body and body-image proportions of \dataset~categories. \dataset~is a balanced demographics dataset, with \textit{talking} being the predominant speaking activity, mainly composed of few people conversations, where audio impaired categories (Speech Impairment and Human Voice Noise) have speakers closer to the camera, and Surveillance Settings has speakers further from it.}
    \label{fig:visual_dataset_characteristics_head_body_graph}
\end{figure*}

\section{Dataset}
\label{sec:dataset}

We propose \dataset, a dataset that aims to show the limitations of current state-of-the-art models by compiling a set of videos from real interactions with varying accessibility of the two key components for ASD: \textit{audio} and \textit{face}. By dividing our dataset into 5 categories with varying degrees of audio and face quality, we can assess how models adapt to these scenarios and which factors are more relevant for ASD. We create a balanced demographics dataset (regarding language, race, and gender), with several challenging factors, complemented with body annotations data. We discuss the process of dataset creation in the following sections.

\begin{table}[!tb]
    \centering
    \small
    \renewcommand{\arraystretch}{1.05}
    \caption{Category feature matrix. Feature description: FA, Face Availability; SO, Speech Overlap;  DS, Delayed Speech; FO, Facial Occlusion; HVB, Human Voice as Background Noise; SS, Surveillance Settings. The absence of a certain feature is presented with $\times$, while its presence with $\checkmark$. Features containing $\mathord{?}$ refer to non-guarantee of its presence or absence. Green cells refer to features favorable for ASD, while red ones are unfavorable.}
    \begin{tabular}{c*{6}{c}}\hline
    
        \makebox[4em]{\textbf{Category}} &
        FA & SO	& DS & FO & HVB & SS  \\
        \hline\hline
        
        Optimal Conditions & 
        \cellcolor{greencell}$\checkmark$ & \cellcolor{greencell}$\times$ & \cellcolor{greencell}$\times$ & \cellcolor{greencell}$\times$ & \cellcolor{greencell}$\times$ & \cellcolor{greencell}$\times$  \\
        
        Speech Impairment & 
        \cellcolor{greencell}$\checkmark$ & \cellcolor{redcell}$\checkmark$ & \cellcolor{redcell}$\checkmark$ & \cellcolor{greencell}$\times$ & \cellcolor{greencell}$\times$ & \cellcolor{greencell}$\times$  \\
        
        Face Occlusion & 
        \cellcolor{greencell}$\checkmark$ & \cellcolor{greencell}$\times$ & \cellcolor{greencell}$\times$  & \cellcolor{redcell}$\checkmark$ & \cellcolor{greencell}$\times$ &  \cellcolor{greencell}$\times$  \\
        
        Human Voice Noise & 
        \cellcolor{greencell}$\checkmark$ & \cellcolor{greencell}$\times$ & \cellcolor{greencell}$\times$ & \cellcolor{greencell}$\times$ & \cellcolor{redcell}$\checkmark$ &  \cellcolor{greencell}$\times$  \\

        Surveillance Settings &
        \cellcolor{redcell}$\mathord{?}$ & \cellcolor{redcell}$\mathord{?}$ & \cellcolor{redcell}$\mathord{?}$ & \cellcolor{redcell}$\mathord{?}$ & \cellcolor{redcell}$\mathord{?}$ &\cellcolor{redcell} $\checkmark$  \\
    
        \hline
        
    \end{tabular}
    \label{table:category-features-dataset}
\end{table}

\subsection{Video and Category Selection}
\label{vid-cat-selection}

We select videos from YouTube and group them into 5 categories based on a set of features, whose values were attributed by human assessment. The main features used for category division are shown in Table~\ref{table:category-features-dataset}, with the complete list in appendix~\ref{sec:categories}. In sum, videos are grouped as follows: 

\begin{itemize}
    \item \textbf{Optimal Conditions}: People talking in an alternate manner, with minor interruptions, cooperative poses, and face availability;
    \item \textbf{Speech Impairment}: Frontal pose subjects either talking via video conference call (\textit{Delayed Speech}) or in a heated discussion, with potential talking overlap (\textit{Speech Overlap}), but ensuring face availability;
    \item \textbf{Face Occlusion}: People talking with at least one of the subjects having partial facial occlusion, while keeping good speech quality (no delayed speech and minor communication overlap);
    \item \textbf{Human Voice Noise}: Communication between speakers where another human voice is playing in the background, with face availability and subject cooperation ensured;
    \item \textbf{Surveillance Settings}: Speaker communication in scenarios of video surveillance, with varying audio and image quality, without any guarantee of face access, speech quality, or subject cooperation.
\end{itemize}

Some important aspects to consider from Table~\ref{table:category-features-dataset}:
1) all categories, aside Surveillance Settings, guarantee face availability, which corresponds to cooperative scenarios and close-up faces; 2) we consider speech delay and overlap as variations of slight speech impairment, thus their grouping in the same category; and 3) Surveillance Settings does not have any guarantee regarding the analyzed features, corresponding to wild conditions. These considerations support the range of ASD difficulty between Optimal Conditions (easier) and Surveillance Settings (harder), since the impairment of audio and face is incremental and controlled throughout the categories. Figure~\ref{fig:categories_dataset} displays representative images of each category and the relative variation of audio and face quality.

\textbf{\dataset~Groups.} Aside category division, we also form two groups of videos for our experiments: \textbf{Easy} and \textbf{Hard}. The easy group contains the categories that more closely resemble AVA-ActiveSpeaker (\textit{Optimal Conditions} and \textit{Speech Impairment}) while the hard group has categories where one or both factors (face and audio) are specifically degraded (remaining 3 categories of \dataset). The inclusion of  Speech Impairment in the easy group relates to how speech overlap is admissible in AVA-ActiveSpeaker (as recurrent from normal conversations) and speech delay as a result of dubbed movies (existent in AVA-ActiveSpeaker).

\subsection{Main Characteristics}

One focus of the proposed dataset is ensuring that each category is balanced regarding language, race, and gender distribution to mitigate any potential bias in future experiments. The languages are grouped into English, European, and Asian, while races are grouped into Caucasian, Afro, and Asian. The considered languages and races, their grouping, and other related considerations are discussed in appendix~\ref{sec:features}. The distribution of demographics, number of speakers, and head-body proportions of \dataset~is presented in Figure~\ref{fig:visual_dataset_characteristics_head_body_graph}. \dataset~only considers two admissible labels, with talking being the dominant speaking activity (contrary to AVA-ActiveSpeaker), and is mainly composed of few people conversations. Surveillance Settings is the one with lesser camera proximity to speakers while Speech Impairment and Human Voice Noise have speakers closer to the camera.

Following the AVA-ActiveSpeaker approach, the maximum length considered for each video is 15 minutes. Contrary to AVA-ActiveSpeaker, where each subvideo duration ranges up to 10 seconds, we segment each subvideo up to 30, with varying video FPS, mainly ranging from 24 to 30. Regarding the number of videos, \dataset~is composed of 164 videos (\textit{vs.} 153 of AVA-ActiveSpeaker), totalling 30 hours of video annotations, divided into train and test with a similar proportion to AVA-ActiveSpeaker (80/20), with each category having roughly the same amount of hours, (\textit{i.e.}, 6 hours) and demographics balance. 

\subsection{\dataset~Annotations}

Body bounding boxes drawing and tracking are obtained using YOLOv5~\cite{redmon2016you} and DeepSort~\cite{wojke2017simple}, serving as input to Alphapose~\cite{fang2017rmpe, li2018crowdpose, xiu2018poseflow}, which outputs pose information for each subject per frame. Then, we obtain face bounding boxes~\cite{roxo2022yinyang} from pose data, using eyes, ears, and nose keypoints as reference for bounding box drawing. The size of face bounding boxes is based on body bounding box height, which is adjusted manually per video to ensure adequate face capture. All face and body annotations are manually revised by a human and adjusted/fully annotated when necessary via Computer Vision Annotation Tool (CVAT)~\cite{boris_sekachev_2020_4009388}. For speaking annotations, we design a custom Graphical User Interface (GUI) program in Python for manual annotation, outputting a file with the format used by AVA-ActiveSpeaker. Further details regarding annotations can be seen in appendixes~\ref{sec:talking_annot} and~\ref{sec:face-detection}.

\section{Experiments}
\label{sec:experiments}

\subsection{Datasets, Models, and Evaluation Metric}
\label{dataset}

\textbf{Datasets.} The AVA-ActiveSpeaker dataset~\cite{roth2020ava} is an audio-visual active speaker dataset from Hollywood movies. With 262 15 minute videos, typically only train and validation sets are used for experiments: 120 for training, and 33 for validation, corresponding to 29,723 and 8,015 video utterances, respectively, ranging from 1 to 10 seconds. The main challenges of this dataset are related to language diversity, FPS variation, the existence of faces with low pixel numbers, blurry images, noisy audio, and dubbed dialogues. Similar to other works, we report the obtained results on the AVA-ActiveSpeaker validation subset. We also use the proposed dataset, \dataset,~which is described in Section~\ref{sec:dataset}. Unless explicitly stated, all models trained in \dataset~use the whole training split (with 5 categories).

\textbf{Models.} The considered models are the ones with state-of-the-art results and publicly available implementations: ASC~\cite{alcazar2020active},  MAAS~\cite{alcazar2021maas}, TalkNet~\cite{tao2021someone}, and ASDNet~\cite{kopuklu2021design}. All models are trained in a two-step process, except TalkNet which is trained end-to-end. MAAS did not provide its Multi-modal Graph Network setup so we present the results from the available implementation.

\textbf{Evaluation Metric.} We use the official ActivityNet evaluation tool~\cite{roth2020ava} that computes mean Average Precision (mAP).

\subsection{Limitations of AVA-ActiveSpeaker Training}

\begin{table}[!tb]
    \centering
    \small
    \renewcommand{\arraystretch}{1.05}
    \caption{Comparison of AVA-ActiveSpeaker trained state-of-the-art models on AVA-ActiveSpeaker and categories of \dataset, using the mAP metric. We train and evaluate each model following the authors' implementation. \textit{OC} refers to Optimal Conditions, \textit{SI} to Speech Impairment, \textit{FO} to Face Occulsion, \textit{HVN} to Human Voice Noise, and \textit{SS} to Surveillance Settings. AVA refers to AVA-ActiveSpeaker.}
    \begin{tabular}{c*{6}{c}}\hline
         \multirow{2}{*}{\textbf{Model}} 

        & \multirow{2}{*}{\textbf{AVA}} & \multicolumn{5}{c}{\textbf{\dataset}} \\

        & & \textbf{OC} & \textbf{SI} & \textbf{FO} & \textbf{HVN} & \textbf{SS}  \\
        \hline\hline
       
        ASC~\cite{alcazar2020active} & 
        83.6 & 86.4 &	84.8  &	69.9  &	66.4 & 51.1 \\
        
        MAAS~\cite{alcazar2021maas} & 
        82.0 & 83.3 & 81.3 & 68.6 & 65.6 & 46.0 \\
        
        TalkNet~\cite{tao2021someone} & 
        91.8 & 91.6 & 93.0 & 86.4 & 77.2 & 64.6 \\
        
        ASDNet~\cite{kopuklu2021design} & 
        91.1 & 91.1 &	90.4 &	78.2 &	74.9 & 48.1 \\
    
        \hline
        
    \end{tabular}
    \label{table:models-performance-dataset-ava}
\end{table}

We start by training models in AVA-ActiveSpeaker and evaluate their performance on AVA-ActiveSpeaker and \dataset, in Table \ref{table:models-performance-dataset-ava}. 

\textbf{Similar to AVA-ActiveSpeaker.}.
Regardless of the model, their performance on Easy categories (Optimal Conditions and Speech Impairment) is similar to the one displayed in AVA-ActiveSpeaker, suggesting the presence of similar characteristics between this group and AVA-ActiveSpeaker. This highlights the importance of face and audio quality for current ASD models, and shows that with high quality data and reliable
face access, simultaneous talk or slight speech delay do not significantly hinder model performance. Furthermore, the similar performance of models in AVA-ActiveSpeaker and Easy categories support the quality of \dataset~annotations. 

\textbf{Face and Audio Importance.}
However, the cross-domain performance is significantly worse in Hard categories. In Face Occlusion, Human Voice Noise, and Surveillance Settings, there is a decrease in performance relative to other categories, suggesting that impairment of face access or audio quality significantly impact models, with a cumulative degrade when both are present (Surveillance Settings). Furthermore, facial occlusion is not as impactful as audio impairment (Human Voice Noise) in ASD, meaning that even when a model can not assess the talking person via face, it can still deduct it via audio analysis. The inverse is not as easily solved, since the existence of audio impairment with human voices (Human Voice Noise) leads to poorer performance relative to the Face Occlusion.

\textbf{The Outlier}. Despite a performance degrade with increasing category difficulty, TalkNet is the best performing model. This could be linked to its end-to-end approach for ASD, contrary to the other models, improving its generalization and performance in cross-domain. Furthermore, TalkNet focuses on long-term temporal context, benefiting from longer videos, which is the case of \dataset.

\subsection{Models Robustness in WASD}

    

\begin{table}[!tb]
    \centering
    \small
    \renewcommand{\arraystretch}{1.05}
    \caption{Comparison of state-of-the-art models on the different categories of \dataset, using the mAP metric. \textit{OC} refers to Optimal Conditions, \textit{SI} to Speech Impairment, \textit{FO} to Face Occulsion, \textit{HVN} to Human Voice Noise, and \textit{SS} to Surveillance Settings.}
    \begin{tabular}{c*{5}{c}}\hline

        \multirow{2}{*}{\makebox[5em]{\textbf{Model}}} &
        \multicolumn{2}{c}{\makebox[5.5em]{\textbf{Easy}}} & 
        \multicolumn{3}{c}{\makebox[5.5em]{\textbf{Hard}}} \\

        & \textbf{OC} & \textbf{SI} & \textbf{FO} & \textbf{HVN} & \textbf{SS}  \\
        \hline\hline
        
        ASC~\cite{alcazar2020active} & 
        91.2 & 92.3 & 87.1 & 66.8 & 72.2 \\

        MAAS~\cite{alcazar2021maas} & 
        90.7 & 92.6 & 87.0 & 67.0 & 76.5 \\

        TalkNet~\cite{tao2021someone} & 
        95.8 & 97.5 & 93.1 & 81.4 & 77.5 \\

        ASDNet~\cite{kopuklu2021design} & 
        96.5 & 97.4 & 92.1 & 77.4 & 77.8 \\
    
        \hline
        
    \end{tabular}
    \label{table:models-performance-dataset}
\end{table}

\begin{figure}
    \centering
    \includegraphics[width=0.49\textwidth]{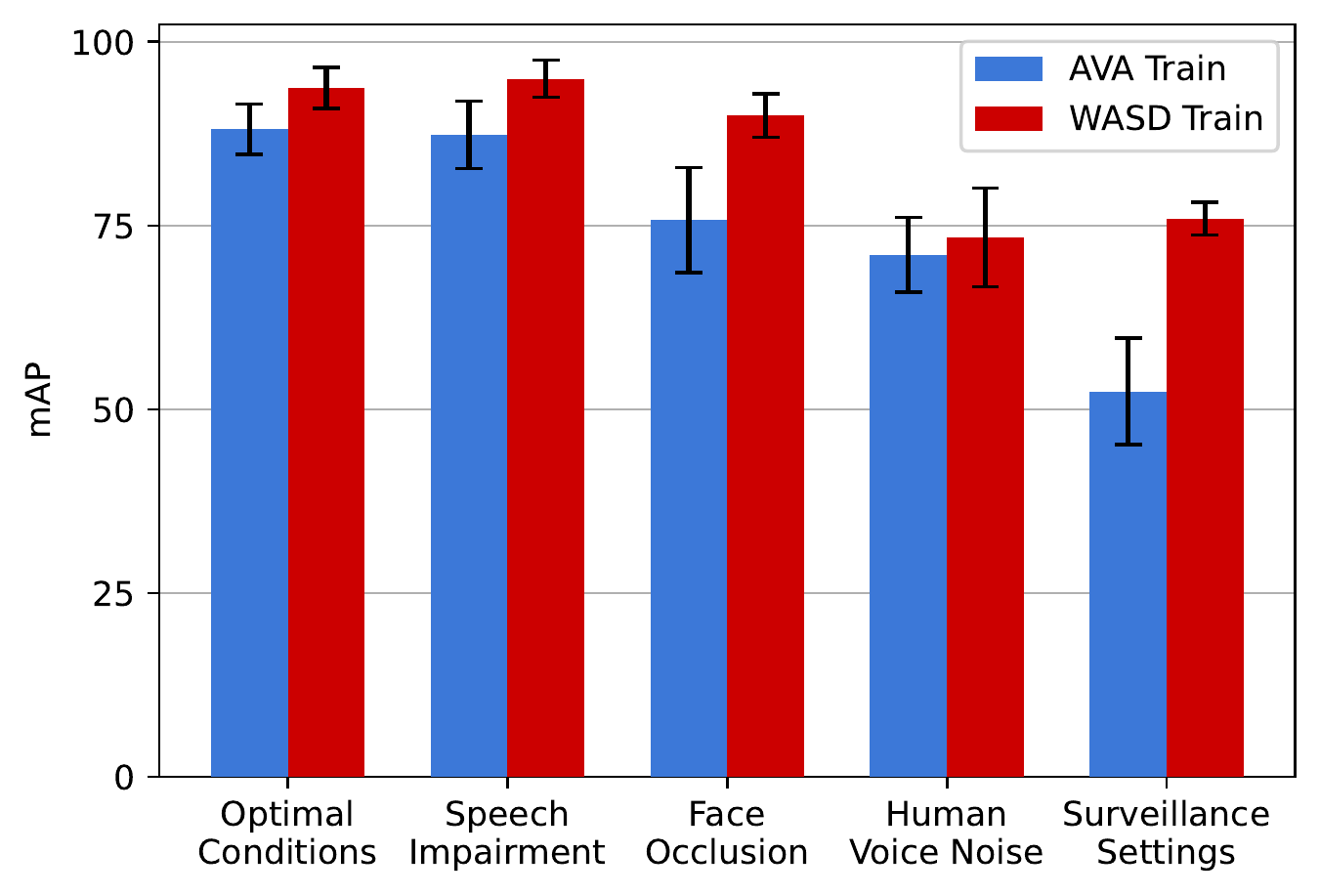}
    \caption{Average performance (mAP) variation of the four models on \dataset{}~categories, when trained on AVA-ActiveSpeaker and \dataset. AVA-ActiveSpeaker is represented as AVA.
    }
    \label{fig:avg_perfomance_variation}
\end{figure}

To evaluate the robustness of models in ASD on challenging data, we train them in \dataset~and compare their performance with AVA-ActiveSpeaker training in Table~\ref{table:models-performance-dataset} and Figure~\ref{fig:avg_perfomance_variation}.

\textbf{Performance Increase}. Relative to AVA-ActiveSpeaker training, models trained in \dataset~tend to slightly improve their performance in Easy setups (Optimal Conditions and Speech Impairment), with higher increase in Face Occlusion and Surveillance Settings scenarios. The increase in Face Occlusion to closer values of those in Easy setups shows that, if trained accordingly, current models can perform ASD in such scenarios. This relates to how models can map different speaker relations in a scene, allowing the inference of one speaker relative to others, even if the face is occluded. Regarding Surveillance Settings, it shows that AVA-ActiveSpeaker does not contain data similar to these settings, but models can perform better in such scenarios if given the proper training. Similar to Face Occlusion, relating different speakers in a scene may give models the tools to perform in such scenarios, even when face access is not reliable.

\begin{figure}[t]
  \centering
    \includegraphics[width=\linewidth]{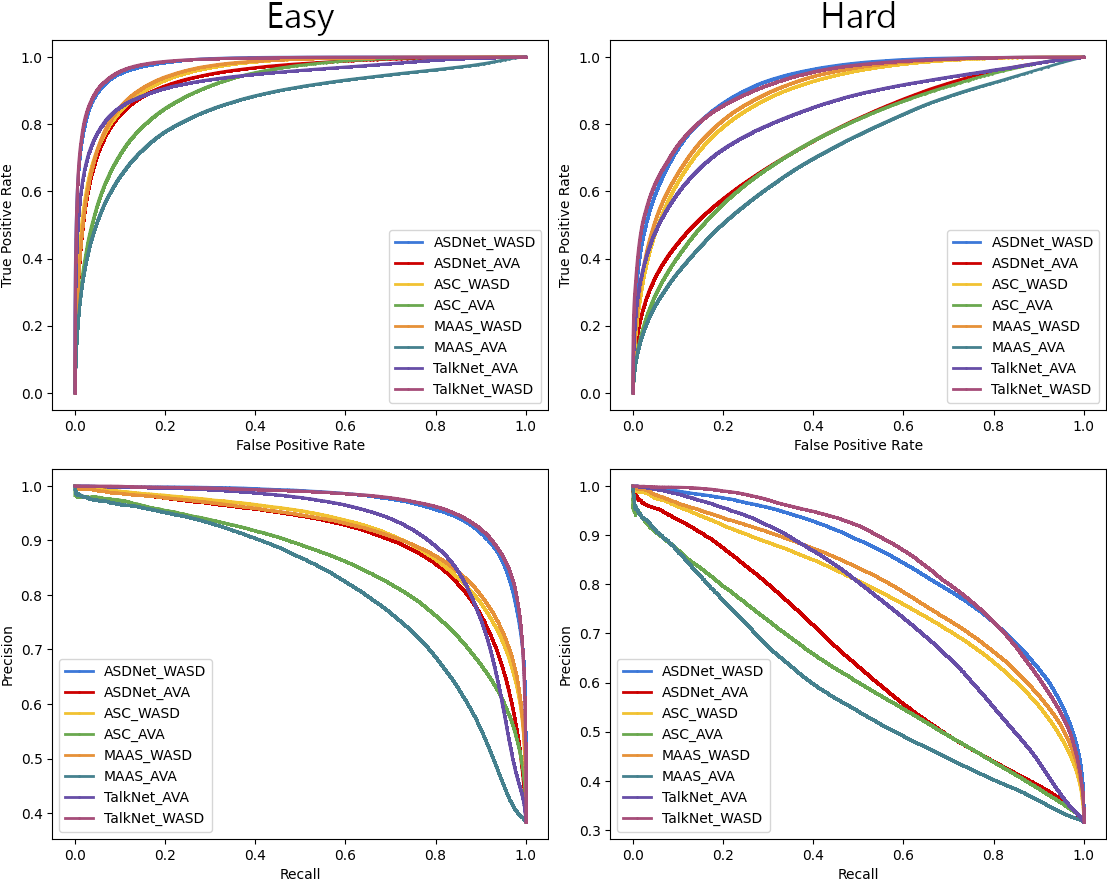}
    
   \caption{ROC and PR curves for models trained in AVA-ActiveSpeaker and \dataset, and evaluated in Easy (left) and Hard (right) groups of \dataset. All models trained in~\dataset~have superior performance to AVA-ActiveSpeaker training. TalkNet trained in AVA-ActiveSpeaker displays a different tendency relative to other models given its long-term analysis and end-to-end training approach. 
   AVA-ActiveSpeaker is represented as AVA.}
   \label{fig:ROC_Precision_Recall}
\end{figure}

\textbf{Model Limitations}. When trained in \dataset, models can not improve their performance in the presence of disruptive/distracting human voice background (Human Voice Noise), which shows the limitations of current approaches. The guaranteed face access may induce a false sense of security to classify a person as talking when they do micro expressions in the presence of (background) human voice. Furthermore, the disparity between the results with human voice background or surveillance settings and the other scenarios (75\% \textit{vs} $>$92\%) shows the limitations of current models to perform in wilder ASD contexts, particularly in impaired audio conditions. 

\textbf{Performance in WASD Groups}. To complement model performance assessment, we compute the Precision-Recall (PR) and Receiver Operating Characteristic (ROC) curves of models in different experimental settings, in Figure~\ref{fig:ROC_Precision_Recall}. 
The results show that:
1) in the Easy group, ASDNet and TalkNet trained in AVA-ActiveSpeaker are competitive with other models trained in \dataset, showing the robustness of the best performing models and the similarity between AVA-ActiveSpeaker and Easy group of \dataset; 2) for the Hard group, all models trained in \dataset~have superior performance relative to AVA-ActiveSpeaker training, suggesting the difference of data between this group and AVA-ActiveSpeaker; and 3) TalkNet trained in AVA-ActiveSpeaker displays a different tendency relative to other models, expressed in both Easy and Hard group, with higher predominance in PR curves. TalkNet has a cautious and precise approach in determining the active speaker (high precision), while not keeping a similar performance in identifying all the active speakers as other models (lower precision with higher recall). This is linked to the lower talking percentage of AVA-ActiveSpeaker and the end-to-end approach of TalkNet with emphasis on long term context: identifying only active speakers with high confidence is a good strategy in AVA-ActiveSpeaker but not as reliable in \dataset.

\begin{figure}[t]
\begin{subfigure}{.49\textwidth}
  \centering
  \includegraphics[width=\linewidth]{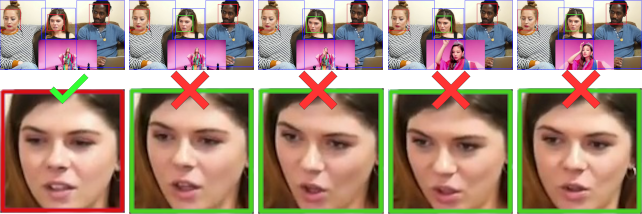}
  \caption{}
  \label{fig:HVBN_error_sequence}
\end{subfigure}
\begin{subfigure}{.49\textwidth}
  \centering
  \includegraphics[width=\linewidth]{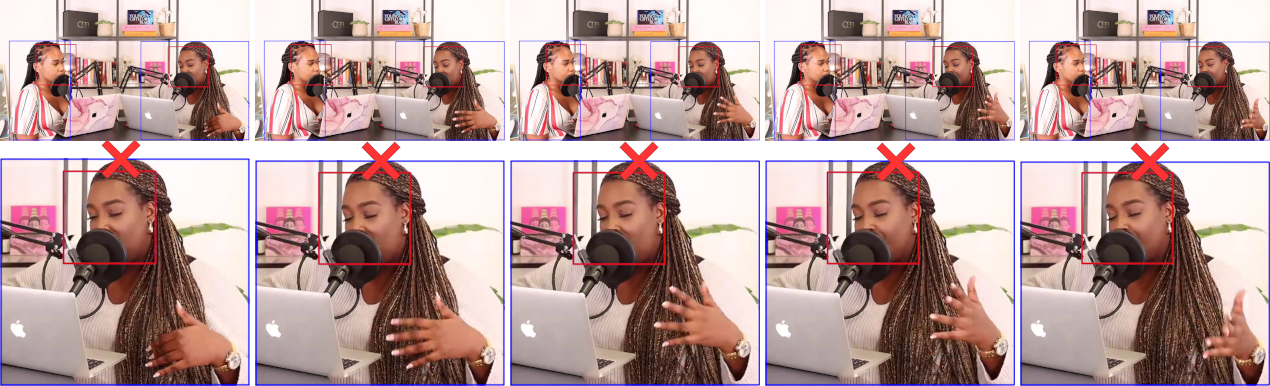}
  \caption{}
  \label{fig:FO_error_sequence}
\end{subfigure}
\begin{subfigure}{.49\textwidth}
  \centering
  \includegraphics[width=\linewidth]{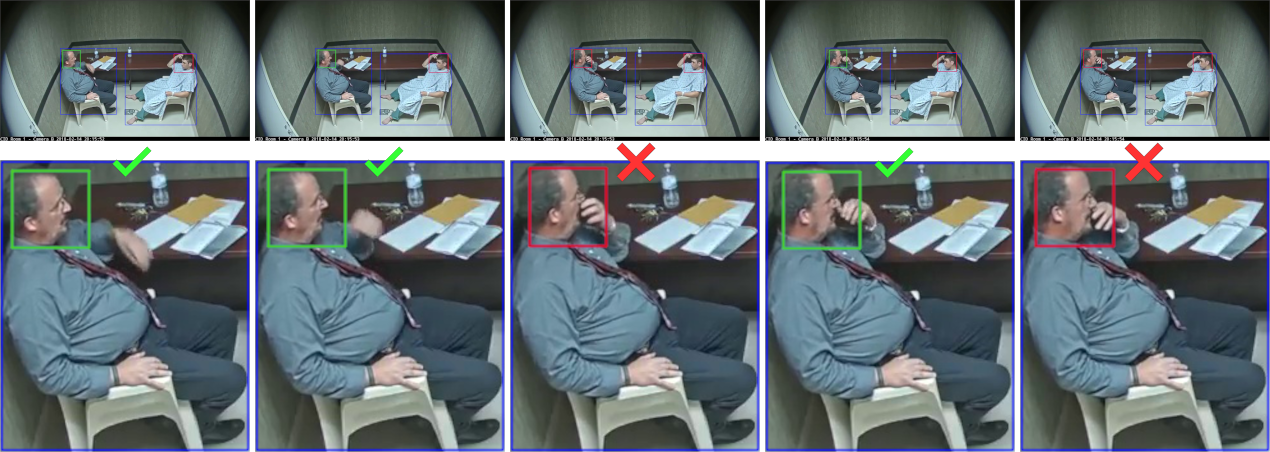}
  \caption{}
  \label{fig:SS_face_occluded_error_sequence}
\end{subfigure}
\caption{Incorrect model inference in different scenarios. Source of misconception: a) awe expression, with sudden and subtle mouth movement, while having human voice in the background; b) partial facial occlusion from scene object; and c)  slight mouth occlusion from hand movement. 
}
\label{fig:3_Error_Sequence_5_small}
\end{figure}

\subsection{Qualitative Analysis}

\begin{figure}[t]
  \centering
    \includegraphics[width=0.49\textwidth]{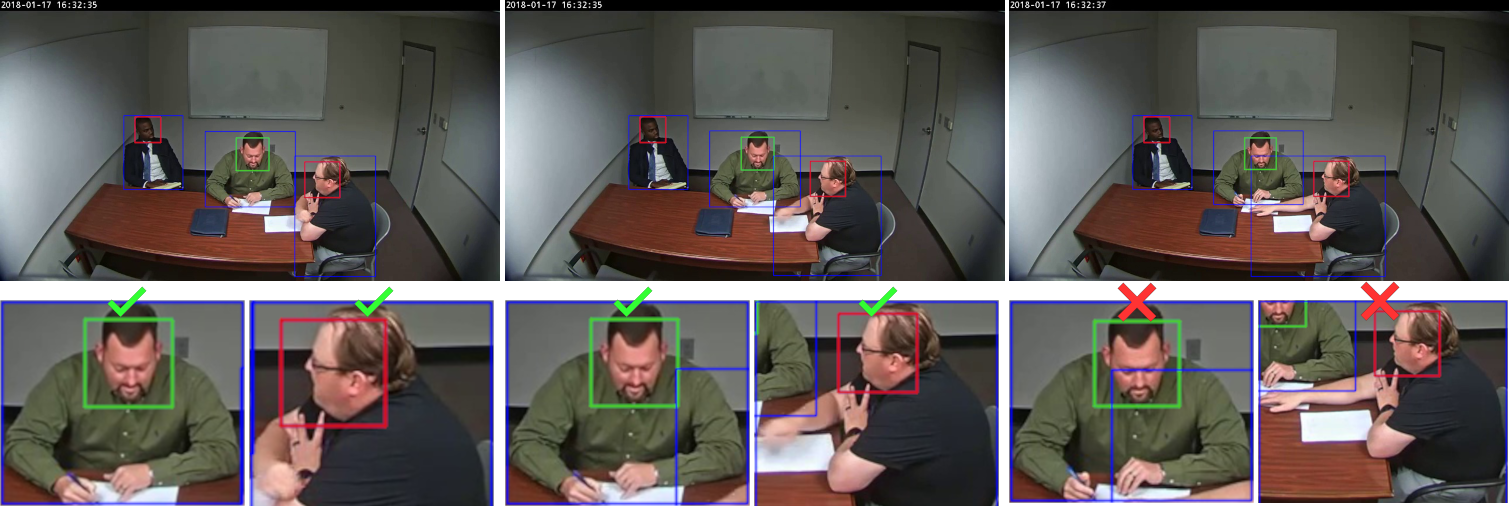}
    
   \caption{Incorrect model inference by mixing active speakers. Hand and arm movement (from right speaker, better viewed with zoom in) suggest a change in conversation between the two speakers, whose analysis would aid understanding speaker swap mid conversation.}
   \label{fig:SS_mismatch_error_sequence}
\end{figure}

We analyze different scenarios where \dataset~is distinctive from AVA-ActiveSpeaker and body data analysis is more relevant for ASD, namely in Human Voice Noise, Face Occlusion, and Surveillance Settings, represented in Figures~\ref{fig:HVBN_error_sequence}, \ref{fig:FO_error_sequence}, \ref{fig:SS_face_occluded_error_sequence}, and \ref{fig:SS_mismatch_error_sequence}, respectively. Head boxes are colored with models predictions, trained in \dataset: green, person is talking; red, not talking. Figures are accompanied with zoom ins containing wrong and correct signs, displaying the correctness of ASD prediction. By not using body information, state-of-the-art models can not reliably deal with scenarios where someone expresses slight lip movement (\textit{e.g.}, awe expression) when another person (not in scene) is talking (Figure~\ref{fig:HVBN_error_sequence}), or with facial occlusion (Figure~\ref{fig:FO_error_sequence}), even in the context of speaker proximity and cooperation. In surveillance settings (Figures~\ref{fig:SS_face_occluded_error_sequence} and \ref{fig:SS_mismatch_error_sequence}) the benefit of body data evaluation is even more pronounced. Accessing hand movement with slight face occlusion helps understanding that the same person is talking (Figure~\ref{fig:SS_face_occluded_error_sequence}), as well as inferring when one person is requesting other to stop talking (Figure~\ref{fig:SS_mismatch_error_sequence}).

    

\section{Conclusion}
\label{sec:conclusion}

We propose \dataset, a challenging ASD dataset with degraded audio quality, facial occlusions, and surveillance conditions. With \dataset~we demonstrate the limitations of state-of-the-art models and AVA-ActiveSpeaker training for wild ASD, particularly in audio impairment and surveillance settings.\dataset~also includes body data annotations to support the development of approaches using body information for wild ASD, given the unreliability of audio quality and subject cooperation in such settings.

\section*{Acknowledgments}

This work was supported in part by the Portuguese FCT/Ministério da Ciência, Tecnologia e Ensino Superior (MCTES) through National Funds and, when applicable, co-funded by EU funds under Project UIDB/50008/2020; in part by the FCT Doctoral Grant 2020.09847.BD and Grant 2021.04905.BD; in part by the C4—Competence Center in Cloud Computing co-financed by the European Regional Development Fund (ERDF) through the Programa Operacional Regional do Centro (Centro 2020), in the scope of the Sistema de Apoio à Investigação Científica e Tecnológica, Programas Integrados de Investigação Científica e esenvolvimento Tecnológico (IC\&DT) under Project CENTRO-01-0145-FEDER-000019


\cleardoublepage

\section*{Appendix}
\renewcommand{\thesection}{\Alph{section}}

\appendix
\section{Talking Annotations}
\label{sec:talking_annot}

\begin{figure}[t]
  \centering
  \includegraphics[width=\linewidth]{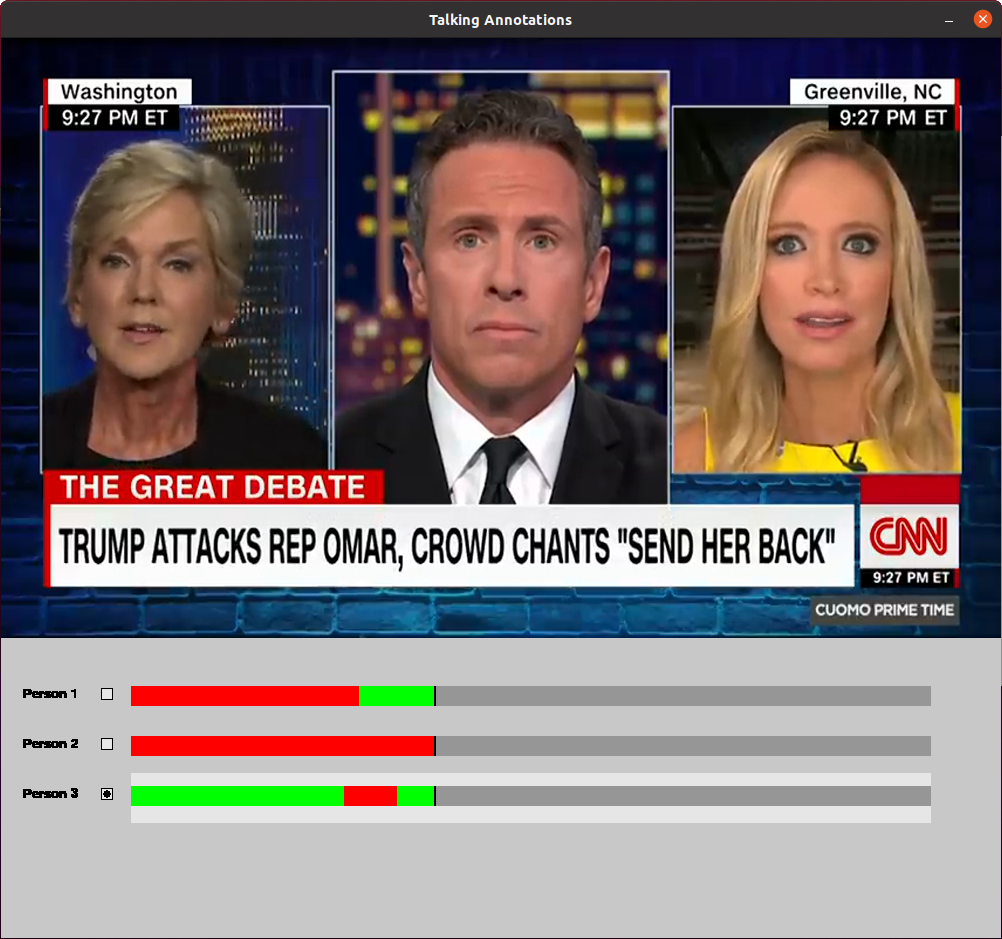}

   \caption{GUI program used for talking annotations. Three speakers are represented by 3 sliders, below the video. Green color refers to speaking, while red represents silence. Leftmost speaker corresponds to the first slider.}
   \label{fig:GUI_example}
\end{figure}

We design a custom Graphical User Interface (GUI) for active speaker annotations, as shown in Figure~\ref{fig:GUI_example}. While the video is running, audio and visual sliders are displayed (one for each speaker), automatically filled with red to denote absence of talking. We select a speaker by pressing the corresponding number key (\textit{e.g.}, third speaker is selected with ``3'' key) and change the speaking label (and slider color) via Ctrl key. To pause, rewind, and forward, we use space, left, and right arrows, respectively. Video time is rewinded and forwarded by 5 seconds with each key press. Prior to GUI launch, we manually set a variable regarding the number of speakers for each video.

\textbf{AVA-ActiveSpeaker Format Conversion}. While performing annotations (either speaking or body/face), we save them on a JavaScript Object Notation (JSON) file (1 file per video), containing all the information used (head/body bounding boxes coordinates, person id, and speaking label), grouped by frame name and person. To convert to AVA-ActiveSpeaker format, we follow the authors guidelines~\cite{roth2020ava}, where each line of the annotation file relates to a time frame of a person in a video. To obtain the time frames, we start at time 0 with increments of video duration per video frames. Each line has the entity id (video name with person id), time frame, face bounding box coordinates, and speaking label. The custom annotations and all AVA-ActiveSpeaker Comma-Separated Value (CSV) files are available at \url{\github}.

\section{Head Bounding Box Annotations}
\label{sec:face-detection}

\begin{figure}[t]
    \centering
    \begin{subfigure}{0.45\textwidth}
      \centering
      \includegraphics[width=\linewidth]{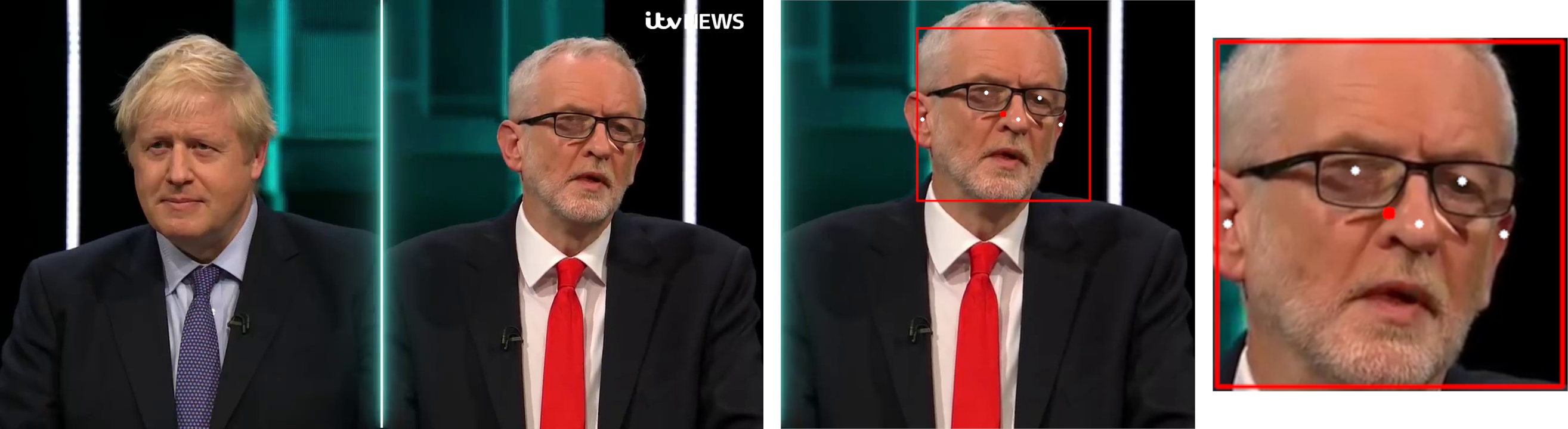}
      \caption{Close frontal.}
      \label{fig:eg_front_close_process}
    \end{subfigure}
    \begin{subfigure}{0.45\textwidth}
      \centering
      \includegraphics[width=\linewidth]{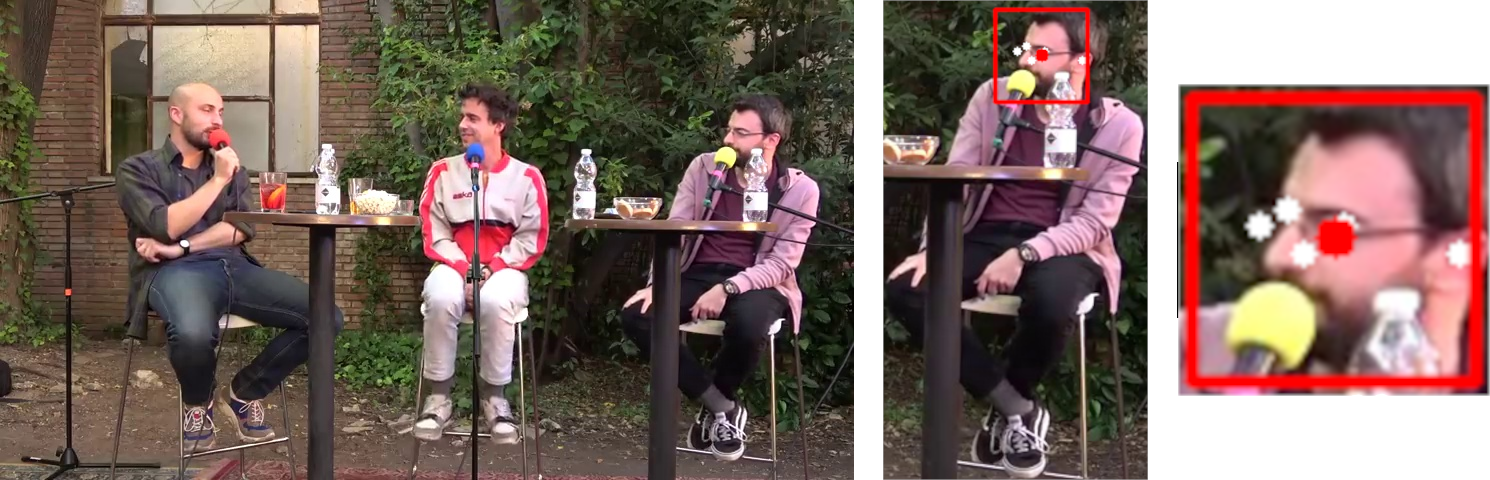}
      \caption{Far frontal.}
      \label{fig:eg_front_far_process}
    \end{subfigure}
    \begin{subfigure}{0.45\textwidth}
      \centering
      \includegraphics[width=\linewidth]{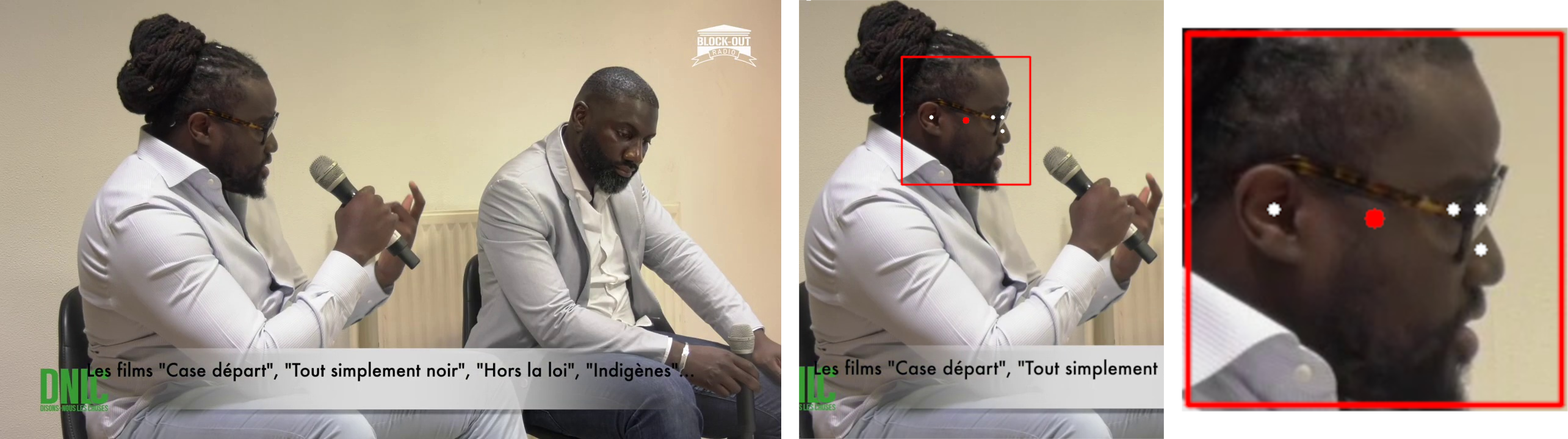}
      \caption{Close sideways.}
      \label{fig:eg_side_close_process}
    \end{subfigure}
    \begin{subfigure}{0.45\textwidth}
      \centering
      \includegraphics[width=\linewidth]{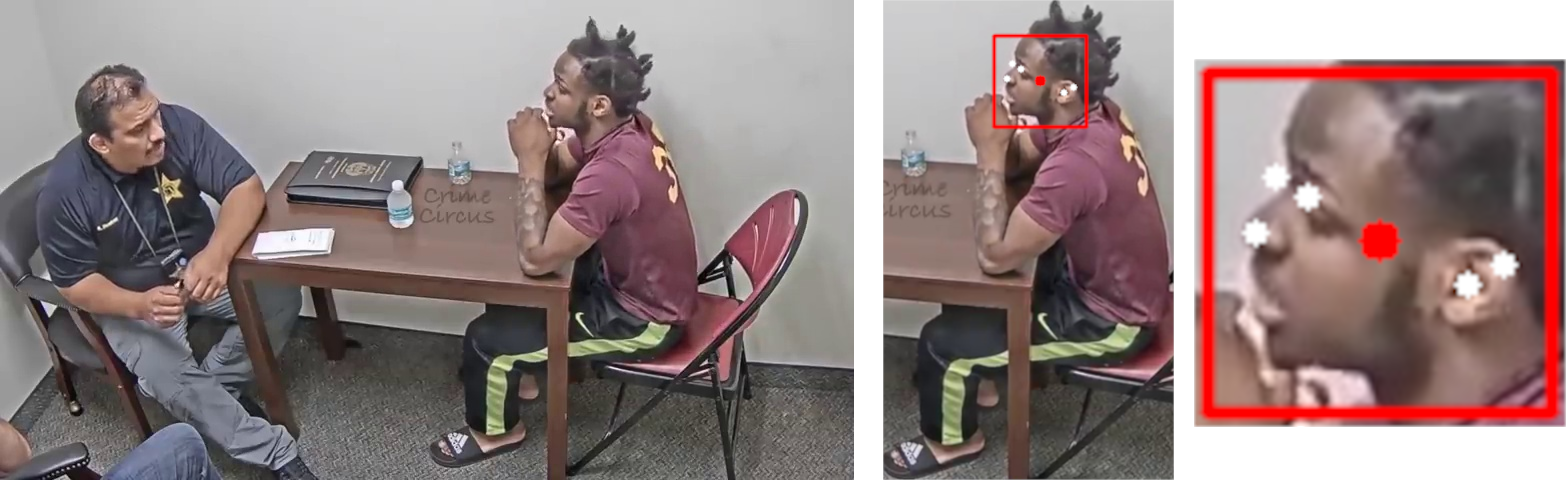}
      \caption{Far sideways.}
      \label{fig:eg_side_far_process}
    \end{subfigure}
    \caption{Head bounding box drawing in different scenarios. From left to right, all images contain the original scenario, head bounding box drawing, and zoom in for better visualization. White dots refer to the reference points used for head bounding box drawing, while the red dot is the head's central point. This approach is suitable for various conditions such as close or far frontal poses ($a$ and $b$), even with facial occlusion, and in side poses at closer or farther positions ($c$ and $d$, respectively).
    }
    \label{fig:Annotations_Examples}
\end{figure}

The default approach for head bounding box annotation is based on pose data. Using Alphapose~\cite{fang2017rmpe, li2018crowdpose, xiu2018poseflow}, we retrieve the $x$ and $y$ coordinates of the right and left ears, right and left eyes, and nose. The head's central point is calculated using the arithmetic mean of the eye-nose reference point (mean of eyes and nose positions) and ears coordinates. Head bounding boxes are centered in the head's central point, with height and width as a fraction of body silhouette height. This fraction is manually set for each video to ensure adequate head area capture. Figure~\ref{fig:Annotations_Examples} displays examples of reference points used and head bounding box drawing in different scenarios. 
In conditions where this approach was not entirely suitable (most Surveillance Settings videos), we annotated manually.  

\section{\dataset~Categories}
\label{sec:categories}

\begin{figure*}[t]
    \begin{subfigure}{\textwidth}
      \centering
      \includegraphics[width=\linewidth]{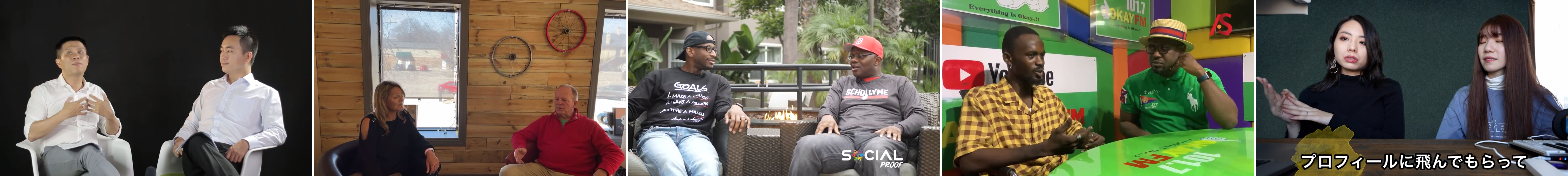}
      \caption{Optimal Conditions.}
      \label{fig:Interview_5_images}
    \end{subfigure}
    \begin{subfigure}{\textwidth}
      \centering
      \includegraphics[width=\linewidth]{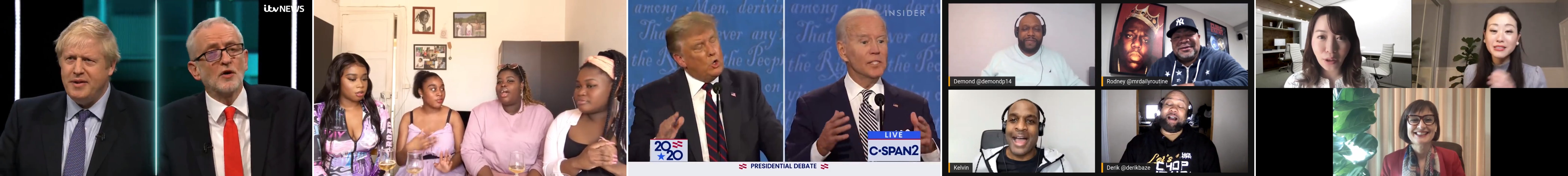}
      \caption{Speech Impairment.}
      \label{fig:Debate_5_images}
    \end{subfigure}
    \begin{subfigure}{\textwidth}
      \centering
      \includegraphics[width=\linewidth]{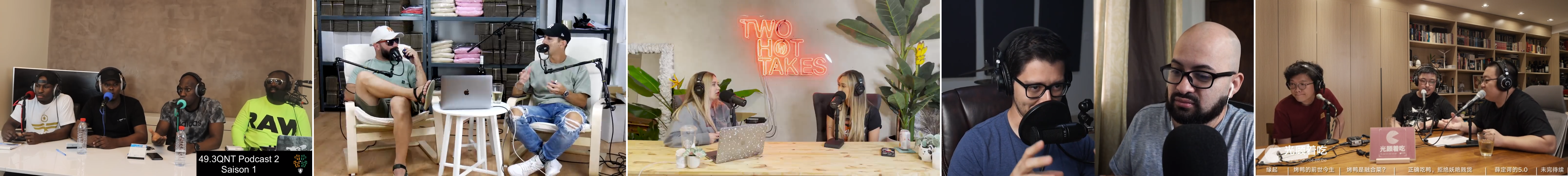}
      \caption{Face Occlusion.}
      \label{fig:Podcast_5_images}
    \end{subfigure}
    \begin{subfigure}{\textwidth}
      \centering
      \includegraphics[width=\linewidth]{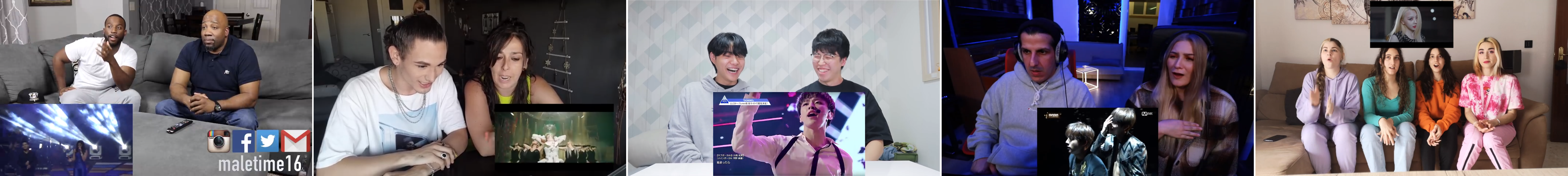}
      \caption{Human Voice Noise.}
      \label{fig:React_5_images}
    \end{subfigure}
    \begin{subfigure}{\textwidth}
      \centering
      \includegraphics[width=\linewidth]{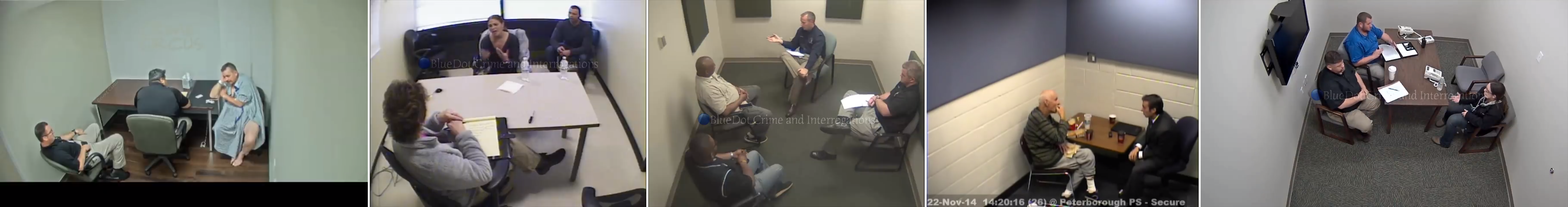}
      \caption{Surveillance Settings.}
      \label{fig:Police_5_images}
    \end{subfigure}
    \caption{Different examples of the considered scenarios for \dataset~categories.
    }
    \label{fig:Categories_Examples}
\end{figure*}

We provide examples of scenarios considered for Wilder Active Speaker Detection (\dataset)~categories in Figure~\ref{fig:Categories_Examples}: \textit{Optimal Conditions} mainly consists of interviews or people talking in an alternate manner, with cooperative poses; \textit{Speech Impairment} refers to political debates, heated discussions, and online interviews/debates; \textit{Face Occlusion} contains various podcast scenarios, where subjects have partial face occlusion from the microphone; \textit{Human Voice Noise} relates to subjects reacting to a video, while it plays in the background, contributing to audio impairment; and \textit{Surveillance Settings} are from indoor surveillance interrogations, with variable audio and image quality (\textit{i.e.}, face access and subject cooperation).

\section{\dataset~Features}
\label{sec:features}

The complete list of all the considered features, and their admissible values, are the following:

\begin{itemize}
    \item \textbf{Facial Occlusion}: Yes or No;
    \item \textbf{Human Voice as Background Noise}: Yes or No;
    \item \textbf{Speech Overlap}: None-Low or Medium-High;
    \item \textbf{Delayed Speech}: Yes or No;
    \item \textbf{Surveillance Settings}: Yes or No;
    \item \textbf{Body Access}: Low, Medium, or High;
    \item \textbf{Audio Quality}: Low or High;
    \item \textbf{Face Availability}: Guaranteed or Non-Guaranteed;
    \item \textbf{Number of People}: from 2 to 7;
    \item \textbf{Number of White People}: from 0 to 5;
    \item \textbf{Number of Afro People}: from 0 to 4;
    \item \textbf{Number of Asian People}: from 0 to 5;
    \item \textbf{Language}: English, European, or Asian;
    \item \textbf{Number of Females}: from 0 to 4;
    \item \textbf{Number of Males}: from 0 to 5;
    \item \textbf{Frames per Second (FPS)}: from 10 to 30;
    \item \textbf{Image Size}: Variable;
    \item \textbf{Video Location}: Indoor or Outdoor;
    \item \textbf{Body-Image Proportion}: Variable;
    \item \textbf{Head-Body Proportion}: Variable;
    \item \textbf{Speaking Percentage}: Variable;
    \item \textbf{Speaking Overlap}: Variable;
    \item \textbf{Luminosity}: Variable.
\end{itemize}

All features with predefined admissible values were attributed by human assessment. Although FPS has various admissible values, they mainly range from 24-30. All other features are continuous values, thus not having a strict set of possibilities: Head-Body Proportion refers to the proportion of head area (bounding box) relative to body area, with similar analogy for Body-Image Proportion; Speaking Percentage and Speaking Overlap is the number of frames with talking and simultaneous talking, respectively; and Luminosity is calculated using the red, green, and blue channels to measure the perceived brightness~\cite{lumino}. Regarding Body Access, we consider three values relating to the visible body area and subject proximity to camera. The CSV containing all the information for each \dataset~video is available at \url{\github}. Regarding the considered languages, we group them as follows:
\begin{itemize}
    \item \textbf{English}: English (USA);
    \item \textbf{European}: Croatian, Dutch, French, German, Italian, Portuguese, Russian, and Spanish;
    \item \textbf{Asian}: Chinese, Japanese, Korean, Pakistanese, and Vietnamese.
\end{itemize}

\end{document}